\documentclass{INTERSPEECH2023}
\interspeechcameraready 

\title{Personalized Predictive ASR for Latency Reduction in Voice Assistants}
\name{Andreas Schwarz$^1$, Di He$^2$, Maarten Van Segbroeck$^2$, Mohammed Hethnawi$^1$, Ariya Rastrow$^2$}
\address{
  $^1$Amazon Alexa, Germany\\
  $^2$Amazon Alexa, USA}
\email{\{asw, deehe, segbrm, hethnm, arastrow\}@amazon.com}

\usepackage{amsmath,graphicx,siunitx}
\usepackage[acronym]{glossaries}
\usepackage{booktabs}
\usepackage{makecell}
\usepackage{tikz}
\usepackage{pgfplots}
\usepackage{listings}
\usepackage{mathtools}
\pgfplotsset{compat=1.17}
\glsdisablehyper

\DeclareMathOperator*{\argmax}{argmax}

\newacronym{AM}{AM}{acoustic model}
\newacronym{ASR}{ASR}{automatic speech recognition}
\newacronym{SLU}{SLU}{spoken language understanding}
\newacronym{TTS}{TTS}{text to speech}
\newacronym{LFR}{LFR}{low frame rate}
\newacronym{LM}{LM}{language model}
\newacronym{EOS}{EOS}{end of sentence}
\newacronym{UPL}{UPL}{user-perceived latency}
\newacronym{OOV}{OOV}{out of vocabulary}
\newacronym{nWER}{nWER}{normalized word error rate}
\newacronym{WERR}{WERR}{word error rate reduction}
\newacronym{WER}{WER}{word error rate}

\newacronym{RNN-T}{RNN-T}{recurrent neural network transducer}
\newacronym{STFT}{STFT}{short-time Fourier transformation}
\newacronym{VAD}{VAD}{voice activity detection}
\newacronym{LSTM}{LSTM}{long short-term memory}
\newacronym{CNN}{CNN}{convolutional neural network}

\begin{document}
\maketitle
\begin{abstract}
Streaming Automatic Speech Recognition (ASR) in voice assistants can utilize prefetching to partially hide the latency of response generation. Prefetching involves passing a preliminary ASR hypothesis to downstream systems in order to prefetch and cache a response. If the final ASR hypothesis after endpoint detection matches the preliminary one, the cached response can be delivered to the user, thus saving latency. In this paper, we extend this idea by introducing predictive automatic speech recognition, where we predict the full utterance from a partially observed utterance, and prefetch the response based on the predicted utterance. We introduce two personalization approaches and investigate the tradeoff between potential latency gains from successful predictions and the cost increase from failed predictions. We evaluate our methods on an internal voice assistant dataset as well as the public SLURP dataset.
\end{abstract}

\section{Introduction}

In voice assistants, a request is processed by multiple systems before the response is ready, starting with \gls{ASR} and the interpretation of the \gls{ASR} hypothesis, and ending with the transmission of the generated \gls{TTS} response to the user, as well as potential execution of external effects such as turning on a smart home device. The steps involved in the generation of the response each contribute varying amounts of latency, adding up to the \gls{UPL}. Besides accuracy, a design goal for voice assistants is minimization of \gls{UPL}. One major contributor to \gls{UPL} is algorithmic latency for utterance endpoint detection \cite{dissecting}. Only when the endpoint of the utterance has been detected with sufficient confidence, e.g., based on acoustic and \gls{ASR} decoder features \cite{maas2018combining,chang2019joint}, the final \gls{ASR} hypothesis can be provided to downstream systems for result generation. Prefetching \cite{chang2020low, bo2021better} has been proposed as a way to reduce \gls{UPL} by already propagating an initial \gls{ASR} hypothesis before the final endpoint has been detected and caching the result. E.g., a second, lower threshold for endpoint detection may be applied to trigger speculative execution using the preliminary \gls{ASR} hypothesis. If the recognition after the final endpoint confirms the preliminary hypothesis, the prefetched response can be returned to the user. This allows hiding a part of the downstream systems' latency within the period between the speculative and the final endpoint. A prerequisite is support for speculative execution in the downstream pipeline, e.g., the possibility of executing the response generation speculatively, while postponing any external effects to after the confirmation by the final endpoint.

\begin{figure}[bt]
    \centering
    \includegraphics[width=\columnwidth]{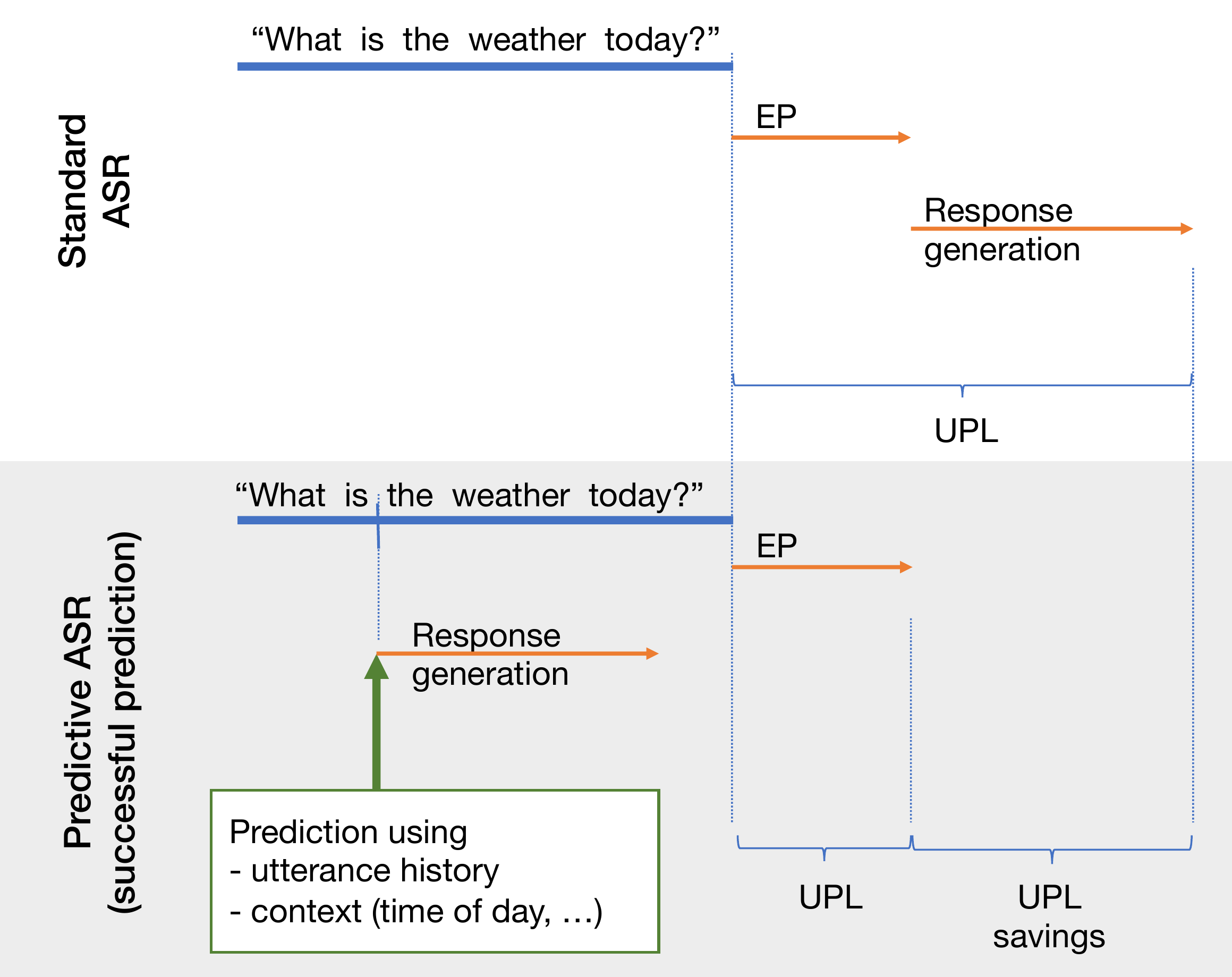}
    \caption{Illustration of major latency contributions for a voice assistant, comparing standard and predictive ASR. User-perceived latency (UPL) is dominated by endpointing (EP) and response generation. Predictive ASR ideally allows hiding the response generation latency within the time between prediction and endpointing, assuming successful prediction.}
    \label{fig:overview}
    \vspace{-2mm}
\end{figure}

In this paper, we propose to extend the concept of prefetching. Instead of using a preliminary endpoint detector to trigger prefetching, we use a prediction strategy that generates a complete utterance from a partially observed utterance while the user is still speaking. This promises a larger window for latency savings compared to endpoint-based prefetching. For queries that can be predicted early in the utterance, we could hide the entire downstream system latency between the time of the initial prediction and the final \gls{ASR} result that matches the prediction. Ideally, the final response playback or intent execution could be carried out immediately after the final \gls{ASR} result is available, assuming the prediction has been correct. Fig.~\ref{fig:overview} illustrates this concept: after the user has said "what is", we can predict that the utterance is going to end in "...the weather today", based on the user's frequent usage of this phrase or additional contextual information such as the time of day. We then use this predicted hypothesis to generate a response. As soon as the endpoint has been detected and the final \gls{ASR} result confirms the prediction, the response can be returned to the user, thus saving latency. In case of a mismatch, the cached response has to be discarded and re-generated, with no effect on latency, but additional compute cost for response generation\footnote{Even in case of mismatch, e.g., the user instead saying "what is the weather today \emph{in Seattle}", a latency benefit may remain due to downstream systems being able to re-use internal caches.}.

In Sec.~\ref{sec:system}, we describe our approach for predictive ASR. In Sec.~\ref{sec:experiments}, we present our experimental implementation and results.

\section{Proposed approach}
\label{sec:system}

We consider a simplified voice assistant system architecture consisting of an utterance endpoint detector, a causal streaming \gls{ASR} model, and a response generator. We make the simplifying assumption that the endpoint detector (EP) and the response generation make up the \gls{UPL} of the system, neglecting \gls{ASR} latency\footnote{An in-depth study of ASR latency contributions can be found in \cite{dissecting}.}:
\begin{equation}
T_\text{UPL} = T_\text{EP} + T_\text{Response},
\end{equation}
where $T_\text{EP}$ is the time between end of the speech and the endpoint decision, and $T_\text{Response}$ is the time that systems downstream from \gls{ASR} require to generate a response and execute the intent. In the literature, prefetching has been proposed for executing response generation and endpoint detection in parallel \cite{chang2020low}, reducing \gls{UPL} in case of success:
\begin{equation}
T_\text{UPL, PF} = \begin{cases*}
\max(T_\text{EP}, T_\text{PF} + T_\text{Response}) & (successful prefetch)\\
T_\text{EP} + T_\text{Response} & (failed prefetch),
\end{cases*}
\end{equation}
where $T_\text{PF}$ is the prefetching latency, which typically is a positive delay between end of speech and the point in time where we trigger prefetching\footnote{Note that some \gls{ASR} models can have a negative partial latency, which would result in a negative delay (see Sec.~\ref{sec:priorwork}).}. We propose to extend prefetching to predict the full utterance transcription before the user has finished speaking the utterance, thus extending the latency saving opportunity. If a correct prediction of the utterance is available by $\Delta T$ before the end of speech in the spoken query, we have a negative prefetching latency, i.e., $T_\text{PF}=-\Delta T$. We call $\Delta T$ the prediction gain.
Premature early prefetching can hinder the efficacy of a predictive ASR strategy. To address this, we propose to implement predictive ASR using combination of a prediction model and a policy that uses a prediction confidence model to accept or reject a predicted transcription. We  describe these models in the following. 

\subsection{Prediction Model}

We consider the task of predicting the full utterance token sequence, $y_\text{full}$, at time $t$ as determining the most probable sequence given all observations available until $t$, $x_t$:
\begin{equation}
\hat y_{\text{full},t} = \argmax_{y_\text{full}} P(y_\text{full}|x_t)    
\end{equation}
Observations would typically consist of the partial utterance audio, but could also include previous interactions and any available contextual information, such as the time of day. For simplification we separate the modeling task into a causal \gls{ASR} model which estimates the partial token sequence up to $t$, $\hat y_t$, and a prediction model which extrapolates this partial token sequence to the most probable complete token sequence:
\begin{equation}
\hat y_t = \argmax_y P(y|x_t)
\label{eq:asr-decoder}
\end{equation}
\begin{equation}
\hat y_{\text{full},t} \approx \argmax_{y_\text{full}} P(y_\text{full}|\hat y_t)
\label{eq:lm-prediction}
\end{equation}
This allows the use of a standard ASR model and decoder to obtain the partial utterance \eqref{eq:asr-decoder} and a standard generative \gls{LM} for the prediction \eqref{eq:lm-prediction}. E.g., prediction could re-use the model and partial computation results from an \gls{LM} which is used for re-scoring \cite{deoras2011fast,kumar2017lattice,xiong2018microsoft}, or leverage a general pre-trained language model such as Generative Pre-trained Transformer (GPT) \cite{radford2018improving}.

\subsection{Prediction Confidence Modeling}

After predicting the full utterance text, a decision needs to be made on whether to propagate the result downstream for prefetching or not. The tradeoff to consider here involves the the probability of success, the cost of a failed speculation attempt, and the latency gain in case of success. While these factors could be modeled in a joint loss function, for now, we only model the prediction confidence, i.e., the probability of the predicted utterance $\hat y_{\text{full},t}$ matching the final ASR hypothesis $\hat y_{\text{full}}$:
\begin{equation}
    P(\hat y_{\text{full},t} = \hat y_{\text{full}}).
\end{equation}
We apply a threshold to this confidence estimate to decide on whether to act on a prediction, and assume we only act on one prediction per utterance.

In the simplest case, this confidence model could be replaced by the probability given by the prediction model. However, training a dedicated confidence model allows the use of additional features which have not been made available to the prediction model, such as the time since the start of the utterance, the confidence of the ASR model for the partial hypothesis, or additional personalized signals.

\subsection{Personalization and Contextualization}

Usage patterns of voice assistants are highly individual and dependent on the context of an interaction. E.g., a user may have the habit of asking for the weather at a certain time of the day, or just use the same consistent phrasing for common requests. Further, requests may depend on information such as the location of the device, entries in a user's playlist, or available smart home devices and their state. E.g., if the partial hypothesis is "turn living room", and the user has a device called "living room light" which is turned off, a completion with "...light on" is very likely. We therefore expect that it is beneficial for prediction accuracy to account for personalization and context. We may condition either the prediction model and/or the confidence model on personal and contextual information such as the recently spoken utterances of a user or the current time of day. While conditioning the prediction model itself is likely to be the most general and powerful approach, practical considerations make the use of personalization in the confidence model attractive as well. E.g., in case an existing, general \gls{LM} is used for prediction, a confidence model could be built with contextual or personalized features and used to select one of several n-best predictions generated from the \gls{LM}. In this paper, we implement personalization in two ways. First, we use the relative frequency of a prediction in a user's recent utterance history as a feature for the confidence model. Second, we augment the predictions from the \gls{LM} with additional predictions obtained by prefix matching in the recent utterance history of a user.

\section{Experiments}
\label{sec:experiments}

We conduct experiments on two English speech datasets. The first is an internal dataset consisting of de-identified user interactions with a voice assistant. This dataset contains four weeks of continuous interaction data from a number of users, with approx. 1700 users in the training partition, and 200 in the development and test partitions, respectively. We use the last three weeks for evaluation, thus allowing us to evaluate the effect of personalized prediction and confidence modeling with at least one week of prior historical context for each utterance under test. The second dataset we experiment on is the SLURP dataset \cite{slurp}, which covers similar domains as our internal dataset, however, consists of artificially generated independent utterances and therefore cannot be used for personalization experiments.

\subsection{Evaluating Prediction Performance}

We distinguish three possible outcomes for an utterance. If the confidence model with a given threshold did not allow any prefetching, there is no impact on latency or cost (\emph{no prefetch}). If prefetching was triggered with a prediction that contains at least one additional word over the partial \gls{ASR} hypothesis, and the prediction turns out to match the final \gls{ASR} hypothesis, we have a potential for a latency gain, without an impact on cost (\emph{successful prefetch}). Finally, if we triggered prefetching with an incorrect prediction, there is no impact on latency, but an impact on cost for repeated response generation (\emph{failed prefetch}). We assume at most one prefetching attempt per utterance.

We first evaluate the rate of successful and failed prefetches relative to the total number of utterances. The rate of successful prefetches corresponds to the fraction of utterances which benefit from prefetching. The rate of failed prefetches corresponds to the relative downstream cost increase due to prefetched responses which need to be discarded and re-generated. To quantify the potential latency gain from prefetching for an utterance, we evaluate the prediction gain $\Delta T$ as the time between the availability of the prediction and the end of speech of an utterance. In other words, this prediction gain corresponds to the extension of the prefetching window achievable over a hypothetical ASR system with perfect endpointing or perfect causal endpoint-based prefetching.

\subsection{System Implementation}
\subsubsection{ASR Decoding}

We use an RNN-T ASR model \cite{rnnt,schwarz21_interspeech} with an 8$\times$1280 \gls{LSTM} encoder, a 2$\times$1280 \gls{LSTM} prediction network, a single-layer joint network, and a total of 148M trainable parameters. The model uses 4k word pieces and is trained on an internal voice assistant dataset.
In order to generate partial ASR hypotheses, we trigger result generation in fixed intervals of 120\,ms. The final ASR result is obtained by decoding the utterance audio after endpoint detection. We note that, while the utterance audio in our dataset contains trailing silence, we compute the prediction gain $\Delta T$ of an utterance as the interval between the last frame used for partial decoding, and the last frame containing speech according to a phonetic alignment of the utterance.

\subsubsection{Prediction}

Our prediction model is a word-level 2-layer \gls{LSTM} \gls{LM} \cite{hochreiter1997long,sundermeyer2012lstm}. It is trained to predict the next token given the previous tokens on a mix of voice assistant utterances and out-of-domain datasets. We use the same type of \gls{LM} in the ASR second-pass-rescoring stage \cite{deoras2011fast,kumar2017lattice,xiong2018microsoft,raju2019scalable} and could therefore theoretically re-use both the model and partially the computations. The model has a the vocabulary size of 283k and 149M trainable parameters (4M in the \gls{LSTM}, the remainder in the input embedding). For evaluation on the SLURP dataset, we train a second model by fine-tuning the first model on data from the SLURP training partition. The perplexity of the models is 15 for the internal dataset test partition, and 31 for the SLURP dataset test partition, respectively (we attribute the difference in perplexity to the fact that the SLURP dataset contains artificial sentences which were generated in written form for research purposes). Prediction is implemented using beam search to generate a set of $N_\text{LM}$ candidate predictions which complete the partial input token sequence until the end-of-sentence symbol. We here chose $N_\text{LM}=4$ as a tradeoff between diversity in prediction candidates and computational effort.

In addition to the prediction from a neural language model, we generate a set of personalized predictions from the past recognized utterances of a user prior to the current utterance (up to 4 weeks of personal usage history). We do this by selecting all previous recognitions which match the prefix of the current partial hypothesis, and combining them with the set of non-personalized predictions obtained from the language model.

\subsubsection{Prediction Confidence Modeling}

We train an ensemble neural classifier (up to 4 dense layers with width 512) to classify whether a prediction matches the final utterance. We train this classifier on predictions from the training partition of the respective dataset and use the development partition for model selection and early stopping. For each utterance, all predictions generated from all partial \gls{ASR} hypotheses are included in the training dataset (excluding predictions with zero predicted tokens), resulting in a training dataset size of 5.5M for our internal dataset and 1.2M for the SLURP dataset.

Our baseline model uses as features the log-probability of the prediction conditioned on the partial hypothesis, and the rank of the prediction in the n-best list, as obtained from the LM. We also use simple text features (number of words and characters in the partial hypothesis and prediction) and the time of the prediction relative to the start of the utterance. Note these features have a minor impact compared to the LM features.

We implement a confidence model which accounts for personalization by adding a personalized feature, which is the log-frequency of the prediction relative to other possible completions of the given prefix from a user's history, with a fallback to -10.0 for utterances not seen in the history.

\subsection{Results and Discussion}

\begin{figure}[t]
    \centering
    \includegraphics[width=\columnwidth]{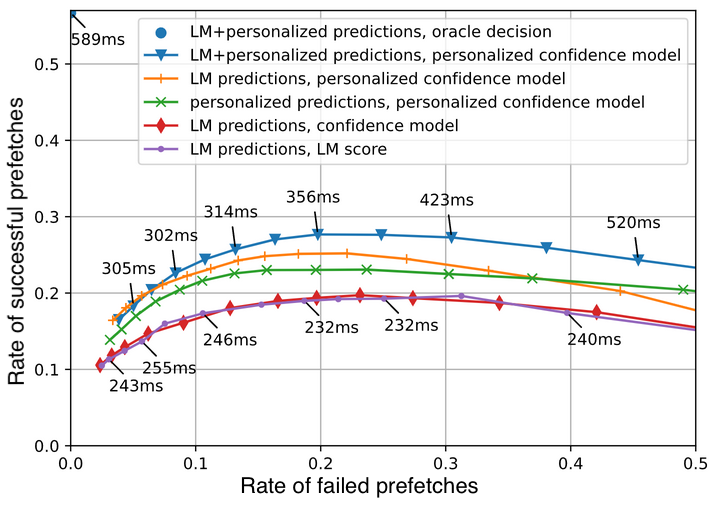}
    \caption{Rate of successful vs. failed prefetches on the internal voice assistant dataset. Text labels show the prediction gain (time between prediction and end of speech) averaged over the successful prefetches.}
    \label{fig:roc-internal}
\end{figure}

\begin{figure}[t]
    \centering
    \includegraphics[width=\columnwidth]{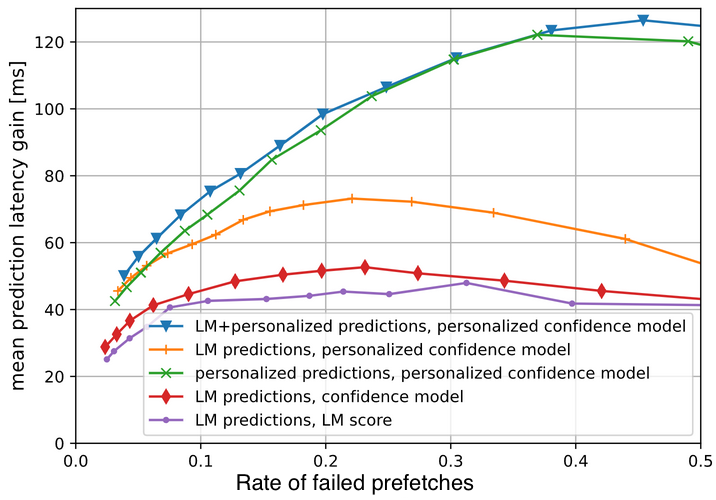}
    \caption{Mean prediction gain over all utterances vs. rate of failed prefetches (internal voice assistant dataset). Rate of failed prefetches corresponds to downstream cost increase.}
    \label{fig:latency-internal}
\end{figure}

Fig.~\ref{fig:roc-internal} shows the tradeoff between successful and failed prefetch rates on our internal dataset (test partition) for varying acceptance thresholds. We also show oracle results, where we accept the first prediction which matches the final recognition; this sets an upper bound of 57\% predictable utterances with our given prediction model. Plot labels display the prediction gain $\Delta T$ after averaging over all successfully predicted utterances. Fig.~\ref{fig:latency-internal} additionally shows the prediction gain averaged over all utterances (with fallback to 0 for utterances with no or failed prefetching attempt), thus reflecting the average potential latency gain while also taking the prefetching rate into account.

We first note that the relationship between successful and failed prefetches is not monotonic, but successful prefetch rate starts decreasing at some point as the acceptance threshold becomes more permissive. This is due to our chosen limitation to at most one prefetch per utterance, which means we waste the potential for prefetching at a more promising point in time if we decide to prefetch too early. On the other hand, we see that more permissive operating points, corresponding to earlier prefetching, lead to a higher latency gain in case of success.

At a maximum, by using a combination of personalized and \gls{LM} predictions, as well as a confidence model which includes personalized features, we can achieve 28\% correctly predicted utterances, at a cost of a 20\% downstream execution overhead (due to failed prefetches). The average prediction gain for successful prefetches at this operating point is 356\,ms, or 1.7 words. Using the same confidence model, but relying merely on \gls{LM} predictions or only personalized predictions reduces the success rate. This indicates that the combination of global and personalized modeling is critical for successful prediction. Finally, using \gls{LM} predictions without any personalized features in the confidence model causes a more significant drop in performance, regardless of whether we use a confidence model or the \gls{LM} score directly as decision criterion for prefetching.

\begin{figure}[b]
    \centering
    \vspace{-3mm}
    \includegraphics[width=\columnwidth]{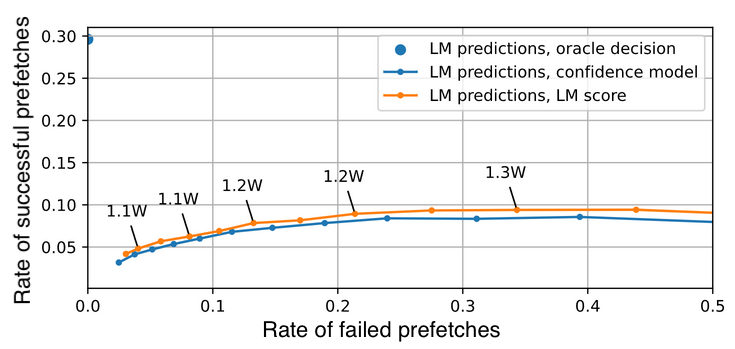}
    \caption{Rate of successful vs. failed prefetches on the SLURP dataset. Text labels show the average number of predicted words for successful prefetches.}
    \label{fig:roc-slurp}
\end{figure}

Fig.~\ref{fig:roc-slurp} shows corresponding results on the SLURP dataset (test partition). For this dataset, we did not estimate the speech endpoint for each utterance, instead we only evaluate prefetching success and failure rates, as well as the average number of predicted words for successful prefetches, which we show in labels. We note that there is no consistent improvement from the use of a confidence model, possibly due to mismatch between the training and test partition distributions, or due to the fact that the confidence model is trained to classify each prediction separately, which is slightly mismatched with our evaluation metric. Overall, prediction accuracy on SLURP is behind our internal dataset, reflecting the higher perplexity which we also observed in the language model. Due to the lack of a per-user history, this dataset also cannot benefit from personalization.

\section{Relation to Prior Work}
\label{sec:priorwork}
In \cite{chang2020low,bo2021better}, prefetching for ASR is described as the process of triggering an early ASR hypothesis generation based on an estimate of the end-of-speech probability. This early hypothesis is used to generate results which are then confirmed or discarded based on the final ASR hypothesis. The authors of \cite{bo2021better} also present observations that streaming \gls{ASR} models are capable of producing partial hypotheses with a negative latency, corresponsing to a prefetch window extension of up to 50ms. Further increase in negative latency could be achieved by applying regularization during model training (FastEmit), although at the cost of degrading \gls{WER} \cite{fastemit}. One major difference in our work is that we use a dedicated prediction model. We therefore do not need to constrain the ASR model training in ways that potentially affect accuracy. Also, prediction using a separate language model allows us to predict up to multiple word tokens of the user utterance. A second difference is that we do not trigger prefetching based on an end-of-utterance probability, but by an estimate of the probability that a prediction matches the final ASR hypothesis.

\section{Conclusion}

We proposed a predictive \gls{ASR} system for voice assistants which prefetches downstream results based on predictions of the full utterance from a partially spoken utterance. The proposed system allows obtaining the correct complete hypothesis on average 300\,ms before the end of speech (i.e., up to 300\,ms latency reduction) for 23\% of the utterances in a voice assistant dataset, while incurring only an 8\% increase in downstream cost due to failed prefetching attempts. The maximum successful prefetch rate of the current system of 28\% might be further increased by lifting the limitation to a single prefetch attempt per utterance, instead allowing multiple parallel or sequential prefetches (incurring higher downstream cost). We further found personalization of predictions to be a critical factor, and expect that an even larger effect could be achieved by conditioning the prediction model itself on the usage history, as well as exploiting additional contextual features such as the time of day or approximate device location.

\clearpage

\bibliographystyle{IEEEtran}
\bibliography{references}
\end{document}